# Augmenting Phrase Table by Employing Lexicons for Pivot-based SMT

**Yiming Cui, Conghui Zhu, Xiaoning Zhu, and Tiejun Zhao**
Harbin Institute of Technology, Harbin, China
`{ymcui,chzhu,xnzhu,tjzhao}@mtlab.hit.edu.cn`

## Abstract

Pivot language is employed as a way to solve the data sparseness problem in machine translation, especially when the data for a particular language pair does not exist. The combination of source-to-pivot and pivot-to-target translation models can induce a new translation model through the pivot language. However, the errors in two models may compound as noise, and still, the combined model may suffer from a serious phrase sparsity problem. In this paper, we directly employ the word lexical model in IBM models as an additional resource to augment pivot phrase table. In addition, we also propose a phrase table pruning method which takes into account both of the source and target phrasal coverage. Experimental result shows that our pruning method significantly outperforms the conventional one, which only considers source side phrasal coverage. Furthermore, by including the entries in the lexicon model, the phrase coverage increased, and we achieved improved results in Chinese-to-Japanese translation using English as pivot language.

## 1 Introduction

The recent improvement in statistical machine translation (SMT) strongly relies on the availability of large parallel data, in order to estimate parameters more precisely. For frequently used language pairs, like English-French, there are large amounts of parallel corpus readily available. However some language pairs, such as English-Dutch, has limited amount of parallel data, due to the popularity of the language pair. In order to solve limitations of parallel data, pivot language method was introduced (de Gispert and Marino, 2006), which bridges two languages through an intermediate language. Furthermore, the pivot language method can also be applied to those popular language pairs, such as Chinese-English, when the bilingual resources are limited for a particular domain.

Phrase pivoting, also called triangulation, is one of the state of the art pivoting method, in which a source-to-target translation model is induced by combining a source-to-pivot and pivot-to-target translation models (Utiyama and Isahara, 2007; Cohn and Lapata, 2007). It aims to obtain a source-to-target model by combining source-to-pivot and pivot-to-target translation models, which has been shown to be generally better than the other pivot methods. However, it has been already reported that the phrase pivoting method may generate a very large phrase table (Utiyama and Isahara, 2007), bringing much noise and losing some phrase pairs if they are not connected to the same pivot phrase (Cui et al., 2013), which will affect the overall translation quality. Moreover, the pivot-based machine translation also suffers from a severe phrase sparsity problem since the loss of translation phrase pairs. Through some simple experiments, we discovered that generally the pivot model is inferior to standard model in phrase coverage, which trained with the same corpus size.

In this paper, we directly employ the word lexical model in IBM models as an additional resource to augment pivot phrase table. Furthermore, we also propose a pruning method which takes into account both of the source and target phrasal coverage, in order to diminish the noise to our phrase table extension method.

## 2 Related Work

Many researchers have investigated pivot language method in statistical machine translation. The first is the sentence translation, also called transfer or cascade method (Khalilov et al., 2008). We first translate the source sentences to pivot language with source-to-pivot translation system, and then translate the pivot language sentences to target language with pivot-to-target translation system. Though very simple, it is time consuming, because it should pass two translation systems consecutively.

The second is the phrase pivoting, also called triangulation method. It combines source-to-pivot and pivot-to-target phrase table to induce a source-to-target phrase table for pivot translation (Cohn and Lapata, 2007; Utiyama and Isahara, 2007; Wu and Wang, 2007). It has been shown that phrase pivoting outperforms other pivoting methods in general (Wu and Wang, 2009).

The third is the synthetic method, also called the pseudo corpus method. It aims to build a source-to-target parallel corpus, using existing source-to-pivot and pivot-to-target corpus (Bertoldi et al., 2008).

Furthermore, Michael Paul and Eiichiro Sumita (2011) investigated the factors that affect the quality of pivot-based machine translation. Zhu et al. (2013a) used random walk method to enhance the connectivity of pivot phrases, creating much more candidates for source phrases and alleviating the OOV problems. Kholy et al. (2013) added connectivity strength features to indicate the quality of alignment projection, and proposed a simple phrase table pruning technique to solve noise problem.

## 3 Phrase Pivoting Method

Given a source-to-pivot and pivot-to-target corpus, we can train two translation models respectively, and induce a pivot translation model, i.e. source-to-target translation model, by combining the source-to-pivot and pivot-to-target translation models through the matched pivot language phrases. In the combining procedure, there are two elements should be taken into account.

The first element is phrase translation probability. Because the source phrases and target phrases are extracted from different corpus, we assume that source phrases are independent with target phrases. When given pivot phrases, we can induce the phrase translation probability $\varphi(s|t)$ as Equation 1.

$$\varphi(s|t) = \sum_{p} \varphi(s|p) \cdot \varphi(p|t) \quad (1)$$

where the $s$, $p$ and $t$ denotes the phrases in the source, pivot, and target respectively.

The second element is lexical weight. We assume $a_1$ and $a_2$ be the alignment information inside phrase pair $(s, p)$ and $(p, t)$ respectively. And we can get the word alignment $a$ of phrase pair $(s, t)$ by the following Equation 2.

$$a = \{(s,t) | \exists p : (s,p) \in a_1 \& (p,t) \in a_2\} \quad (2)$$

Then we can estimate the lexical translation probability $w(s|t)$ by induced word alignment information.

$$w(s|t) = \frac{count(s,t)}{\sum_{s'} count(s',t)} \quad (3)$$

## 4 Approach

As we mentioned above, the pivot-based machine translation suffers from serious noise problem caused by the compounded error in each translation model. In addition, the induced pivot translation model could be order of magnitude larger than unmerged models, which is almost occupied by noise phrase pairs. We tackle this problem by first pruning the enlarged translation model in order to exclude noise phrase pairs. Secondly, we augment translation model by employing additional entries from lexicon models.

### 4.1 Modified Top-N Pruning Method

In pivot-based machine translation, a basic phrase table pruning technique is the top-N pruning. That is to say, we select the top-N candidates for a given source phrase in pivot phrase table. It is proved to be a simple and useful phrase table pruning strategy (Zhu et al., 2013a; Kholy et al., 2013).

However, we have noticed that most of the pruning methods seldom take the source phrase coverage into account, and we believe that the noise not only exists in target phrases, but also in source phrases. Motivated by this, we propose a modified top-N pruning method, which will lead to a much compact phrase table, and significant improvements in translation performance.

When given a source phrase *S*, we select its top-N scored phrases, and discard those who are not in the top-N list. To select these top-N candidates, based on the log-linear model, we multiply each feature (such as translation probabilities and lexical weights) by its optimized decoding weight, which are computed in the tuning procedure. Formally, when a phrase pair (*S,T*) is given, its score can be calculated as Equation 4.

$$score(S,T) = \sum_{i=1}^{n} W_i \cdot \log(F_i) \tag{4}$$

where $W_i$ is the *i*th optimized decoding weight of (*S,T*), $F_i$ is the *i*th feature of (*S,T*), and *n* represents the number of features.

Through above steps, we have done the basic top-N pruning. That is to say, we prunes phrases with respect to the source side of the phrase pairs, so that only *N* phrases for each source phrase are preserved.

Then we further applied similar pruning method to the target phrases. We performed top-M pruning with respect to the target side of the phrase pairs after top-N pruning, so that only *M* phrases for each target phrase are kept as translation candidates. Note that, some attention should be paid to the selection of pruning threshold *N* and *M*. We will have a further discussion about this in Section 5.

### 4.2 Augmenting Phrase Table by Adding Lexicons

As we mentioned in Section 1, the pivot-based machine translation suffers from serious phrase sparsity problem, and even a basic translation unit, unigrams, may not be covered after inducing phrase pairs of pivot model.

From a macro perspective, this is because the pivot method cannot make full use of the source-to-pivot and pivot-to-target corpus resources. When we combine the source-to-pivot and pivot-to-target phrase table, those phrases which are not connected through the same pivot phrases will be discarded, even though similar phrases may be matched in the pivot side. However, after we look into this problem, we found that the phrase coverage problem might result from word alignment error, which might prevent us from extracting useful phrase pairs.

To solve this problem, we introduce a novel approach to enrich the phrase pairs in pivot phrase table, in order to increase the phrase coverage in pivot-based machine translation and improve overall translation performance.

Besides the phrase-based translation model, we directly employ the word lexical model in IBM models (Brown et al., 1993) as an additional resource to extend pivot phrase table. After the word alignment is done, we can get a lexical model, which contains word-to-word translation table with its conditional translation probability (word-based model). So a source-to-target word lexical model can be rebuilt in a similar way to induce a pivot phrase table, which is to combine source-to-pivot and pivot-to-target lexical models.

Formally, given a source-to-pivot word pair <*s,p*> and pivot-to-target word pair <*p,t*>, if the pivot word is the same, we can get a source-to-target word pair <*s,t*> with its conditional probabilities calculated in Equation 5 and 6.

$$\psi(s|t) = \sum_{p} \psi(s|p) \cdot \psi(p|t) \tag{5}$$

$$\psi(t|s) = \sum_{p} \psi(p|s) \cdot \psi(t|p) \tag{6}$$

where $\psi(s|t)$ and $\psi(t|s)$ denotes the direct and inverse conditional probability respectively. Note that, it is possible to project a *NULL* word to a target word in word alignment, but here we say that the word pairs such as <*s,NULL*> and <*NULL,t*> are not participated in generating the pivot lexical model. Because they cannot be used to produce reliable source-to-target word pairs.

To formulate a phrase table, there are two extra things should be calculated, that is lexical weight. In this paper, we just copy the corresponding translation probabilities, as shown in Equation 7 and 8. We also tried conventional lexical weight calculation method (similar to Equation 3), as well as constant values for lexical weights. The results and further discussions are shown in Section 5.

$$lex(s|t) = \psi(s|t) \tag{7}$$

$$lex(t \mid s) = \psi(t \mid s) \tag{8}$$

Until now, we can combine the lexical translation table and original pivot phrase table to form a new pivot phrase table. Given a word pair <s,t> in lexical translation table, we just check if it is in the original pivot phrase table. If exists, we do not add <s,t> into it, or we just add this item into the original pivot phrase table.

## 5 Experiments

### 5.1 Experiment Setup

In this paper, we build a Chinese-Japanese translation system without using parallel corpus. We select English as pivot language, because of its large availability of bilingual corpus. We randomly select 50K Chinese-English and 59K Japanese-English sentences from an in-house parallel corpus respectively, containing spoken corpus of various domains.

| System | N | M | BLEU | Size | Percentage |
|---|---|---|---|---|---|
| Baseline (Top100) | 100 | N/A | 20.68 | 8851K | 100% |
| +Top50 | 50 | N/A | **22.02** | **5343K** | **60.4%** |
| +Top20 | 20 | N/A | **21.84** | **2783K** | **31.4%** |
| +Inv50 | N/A | 50 | **21.82** | **3072K** | **34.7%** |
| +Top50+Inv100 | 50 | 100 | **22.47** | **2763K** | **31.2%** |
| +Top50+Inv50 | 50 | 50 | **21.90** | **2301K** | **25.9%** |
| +Top50+Inv20 | 50 | 20 | 20.98 | 1687K | 19.1% |

Table 1. Different N, M threshold settings for modified top-N phrase table pruning

We used Chinese-Japanese parallel corpus to make a 2K tuning set and 1K test set with single reference each. Note that the training set, tuning set and test set are independent each other.

We carry out our experiments using an open-source phrase-based SMT toolkit Moses (Koehn et al., 2007). Word alignment and phrase extraction are done by GIZA++ (Och and Ney, 2003). 5-gram language models are trained by SRILM toolkit (Stolcke, 2002). We use MERT (minimum error training) for parameter tuning (Och, 2003). Because of MERT's instability, we tune every translation system 5 times independently and take the average BLEU score (Clark et al., 2011; Zhu et al., 2013b). The translation quality evaluation is done by case-insensitive BLEU-4 metric (Papineni et al., 2002). The statistical significance test is also carried out with paired bootstrap resampling method (Koehn, 2004).

### 5.2 Experiment of Modified Top-N Pruning

We applied several experiments to show the performance of modified top-N pruning method. As illustrated in Section 4.1, it is hard to choose proper values for *N* and *M* (represents direct/inverse pruning threshold respectively). So we carried out a series of experiments to try various combinations of *N* and *M* values. The experiment results are shown in Table 1. Because the size of the original pivot phrase table is huge (about 252M), we set a pruned phrase table with top-100 as our baseline system. The values in the *Size* column represent the number of phrase pairs. *Inv* represents inverse top-N pruning, which is the second step of modified top-N pruning method. The BLEU scores that statistically significant than the baseline (above 95% level) are marked with bold face. From Table 1, we observed that:

(1) Top-N pruning proved to be a simple and useful phrase table pruning method. The pruned phrase table is much more compact than original one. As the noise in phrase table has been removed, the translation performance is also improved.
(2) Modified top-N pruning method generally outperform the traditional one. The scale of best system (+Top50+Inv100) is reduced to 31.2% of baseline system, and BLEU score raised 1.79 points. The best system also exceed the conventional top-N pruning method, both in scale and translation performance. When compared to the original pivot phrase table, the scale of best system significantly reduced to 0.01%, which will save much storage space and training time.

(3) With the growing of *Inv* (see row 5 to 7), the translation performance also improved, but the scale of phrase table slightly increased. This is because, the higher we set an *M* value, the more we get source phrase candidates.

Compared to *M* value, it is easy to set *N* value. Because for one source phrase, the number of target phrase candidates cannot be infinitively growing, even if the training corpus increases. So we empirically conclude that the *N* value should not exceed 100 in most cases.

On the contrary, the inverse pruning threshold *M* should not set to a small value, in case of losing diversity of source phrases and increasing the OOVs. As we can see in Table 1, when the *M* value decreases to 20, the BLEU score drops dramatically. So we suggest that, for most circumstances, the *M* value should not be less than 100. We will investigate how to automatically optimize pruning thresholds for a given pivot-based translation system in the future.

### 5.3 Experiment of Augmenting Phrase Table

In this experiment, we will test the performance of the phrase table augmenting. As we mentioned in Section 4.2, we try different lexical weight calculation methods to test their performances. To make the results more precise, we use top-N pruning method to lexical translation table, and only 20 candidates for each source word are preserved. Finally, we obtain a 426K lexical translation table.

| System | LEX | BLEU | OOV |
|---|---|---|---|
| Baseline_Top100 | N/A | 20.68 | 400 |
| Baseline_OPT | N/A | 22.47 | 400 |
| +re-estimate | re-est | **22.68** | **191** |
| +copy | copy | **22.53** | **191** |
| +constant | $e^{-10}$ | **22.89** | **191** |

Table 2. Performance of phrase augmentation

In Table 2, the *LEX* column shows the method to calculate lexical weight. The *re-estimate* means conventional lexical weight calculation method, and *copy* means the approach we mentioned in Section 4.2, i.e. copy corresponding probabilities, and $e^{-10}$ means constant value $e^{-10}$. *Baseline_OPT* represents the best system we obtained in Section 5.2 (with N=50, M=100). Besides the BLEU metric, we also show the number of OOVs, which indicates the performance of phrase coverage.

As we can see that, after applying phrase table augmenting (not matter how we calculate lexical weights), these systems still outperform the baseline (Baseline_Top100) significantly. Furthermore, it is worth mentioning that as the phrase coverage improved by adding lexical translation table, the number of OOVs are significantly reduced from 400 to 191 (47.75% of its origin).

Here, our experimental results can be summarized as follows:

(1) The system performance changes slightly with different lexical weight calculation method (see row 3 to 5). One of the reasons is that the number of lexical phrases added to our phrase table is far smaller than the induced phrase table, and thus we observe little impact to the system.
(2) The phrase extended systems (see row 3 to 5) are not significantly better than *Baseline_OPT* in BLEU scores (only improved 0.06~0.42). In our experiments, we used a test set consisting only single reference, which may potentially underestimate the gains by reducing OOVs. To verify this, we analyzed the rank of translation candidates during decoding. We randomly selected some examples of negative results in Table 3.

| Test word | Baseline system | Our system | Testset |
|---|---|---|---|
| 汽水 | サイダー | ソーダ | サイダー |
| 古典 | 古典 | クラシック | 古典 |

Table 3. Negative results analysis

In the test sentence "我想喝一杯汽水" (I want a glass of soda.), the test word "汽水" (soda) is translated the same in the baseline and test set, but our system result "ソーダ" is another representa-tion of test word "汽水", so as another case "古典" (classical). If we use a multi-referenced test set, the BLEU score may raise up, due to the test set covered with various synonyms. However, on the other side, the phrase coverage problem relived (OOV reduced) by our method, with minor phrases translated to its synonyms, is acceptable in practical use. Furthermore, sometimes it is much more meaningful to cover more phrases than a slight improvement in BLEU score, which will convey more informations.

# 6   Conclusion and Future Work

In this paper, we presented a modified phrase table pruning method, which takes both of the source and target phrasal coverage into account, to alleviate the noise problem in pivot-based machine translation. In addition, we also introduced a novel approach to improve phrase coverage in phrase table, via employing the word lexical model in IBM models as an additional resource. Experiment results show that our pruning method significantly outperforms the original one, not only in reducing the size of phrase table, but also in improving overall translation performance. And after applying our phrase table extension strategy, the phrase coverage increased, in terms of OOV reduced to under 50% of its origin.

In our future work, we are planning to integrate linguistic information to the phrase table augmenting method, in order to obtain a better pivot phrase table. Furthermore, we are also going to investigate new way to integrate lexical model into pivot model, such as linear interpolation etc.

## References

Nicola Bertoldi, Madalina Barbaiani, Marcello Federico, and Roldano Cattoni. 2008. Phrase-based Statistical Machine Translation with Pivot Languages. In *Proceedings of the International Workshop on Spoken Language Translation (IWSLT)*, pages 143-149.

Peter F. Brown, Vincent J. Della Pietra, Stephen A. Della Pietra, Robert L. Mercer. 1993. The mathematics of statistical machine translation: Parameter estimation. In *Computational linguistics*, volume 19(2), pages 263-311.

Yiming Cui, Conghui Zhu, Xiaoning Zhu, Tiejun Zhao and Dequan Zheng. 2013. Phrase Table Combination Deficiency Analyses in Pivot-based SMT. In *Proceedings of 18th International Conference on Application of Natural Language to Information Systems (NLDB)*, pages 355-358.

Tevor Cohn and Mirella Lapata. 2007. Machine Translation by Triangulation: Making Effective Use of Multi-Parallel Corpora. In *Proceedings of the 45th Annual Meeting of the Association for Computational Linguisitics (ACL)*, pages 348-355.

Jonathon H. Clark, Chris Dyer, Alon Lavie, and Noah A.Smith. 2011. Better hypothesis testing for statistical machine translation: controlling for optimizer instability. In *Proceedings of the 49th Annual Meeting of the Association for Computational Linguistics: Human Language Technologies: short papers*, pages 176-181.

Adria de Gispert and Jose B. Marino. 2006. Catalan-English statistical machine translation without parallel corpus: bridging through Spanish. In *Proceedings of 5th International Conference on Language Resources and Evaluation (LREC)*, pages 65-68.

Philipp Koehn, Hieu Hoang, Alexandra Birch, Chris Callison-Burch, Marcello Federico, Nicola Bertoldi, Brooke Cowan, Wade Shen, Christine Moran, Richard Zens, Chris dyer, Ondřej Bojar, Alexandra Constantin, and Evan Herbst. 2007. Moses: open source toolkit for statistical machine translation. In *Proceedings of the 45th Annual Meeting of the Association for Computational Linguistics (ACL) on Interactive Poster and Demonstration Sessions*, pages 177-180.

Philipp Koehn, Franz Josef Och, and Daniel Marcu. 2003. Statistical phrase-based translation. In *Proceedings of Human Language Technology conference of the North American chapter of the Association for Computational Linguistics (HLT-NAALC)*, pages 127-133.

Philipp Koehn. 2004. Statistical significance tests for machine translation evaluation. In *Proceedings of the Empirical Methods in Natural Language Processing Conference (EMNLP'04)*, pages 388-395.

M. Khalilov, Marta R. Costa-juss, Jos A. R. Fonollosa, Rafael E. Banchs, B. Chen, M. Zhang, A. Aw, H. Li, Jos B. Mario, Adolfo Hernndez, and Carlos A. Henrquez Q. 2008. The TALP & I2R SMT Systems for IWSLT 2008. In *Proceedings of the International Workshop on Spoken Language Translation (IWSLT)*, pages 116–123.

Ahmed El Kholy, Nizar Habash, Gregor Leusch, and Evgeny Matusov. 2013. Language Independent Connectivity Strength Features for Phrase Pivot Statistical Machine Translation. In *Proceedings of the 51st Annual Meeting of the Association for Computational Linguistics (ACL)*, pages 412-418.

Franz Josef Och and Hermann Ney. 2003. A Systematic Comparison of Various Statistical Alignment Models. In *Computational Linguistics*, volume 29(1), pages 19-52.

Franz Josef Och. 2003. Minimum error rate training in statistical machine translation. In *Proceedings of the 41st annual Meeting of Association for Computational Linguistics (ACL)*, pages 160-167.

Michael Paul and Eiichiro Sumita. 2011. Translation Quality Indicators for Pivot-based Statistical MT. In *Proceedings of 5th International Joint Conference on Natural Language Processing (IJCNLP)*, pages 811-818.

Kishore Papineni, Salim Roukos, Todd Ward, and Weijing Zhu. 2002. BLEU: a method for automatic evaluation of machine translation. In *Proceedings of 40th Annual Meeting of the Association for Computational Linguistics (ACL)*, pages 311-318.

Andreas Stolcke. 2002. SRILM - an Extensible Language Modeling Toolkit. In *Proceedings of the International Conference on Spoken Language Processing (ICSLP)*, volume 2, pages 901-904.

Masao Utiyama and Hitoshi Isahara. 2007. A Comparison of pivot methods for phrase-based statistical machine translation. In *Proceedings of Human Language Technologies (HLT)*, pages 484-491.

Hua Wu and Haifeng Wang. 2009. Revisiting Pivot Language Approach for Machine Translation. In *Proceedings of the 47th Annual Meeting of Association for Computational Linguistics and the 4th International Joint Conference of Natural Language Processing of the AFNLP*, pages 154-162.

Hua Wu and Haifeng Wang. 2007. Pivot Language Approach for Phrase-based Statistical Machine Translation. In *Proceedings of 45th Annual Meeting of the Association for Computational Linguistics (ACL)*, pages 856-863.

Xiaoning Zhu, Zhongjun He, Hua Wu, Haifeng Wang, Conghui Zhu, and Tiejun Zhao. 2013a. Improving Pivot-based statistical machine translation using random walk. In *Proceedings of the conference on Empirical Methods in Natural Language Processing (EMNLP)*, pages 524-534.

Conghui Zhu, Taro Watanabe, Eiichiro Sumita, and Tiejun Zhao. 2013b. Hierarchical Phrase Table Combination for Machine Translation. In *Proceedings of 51st Annual Meeting of the Association for Computational Linguistics (ACL)*, pages 802-810.